\documentclass[10pt,twocolumn,letterpaper]{article}

\usepackage{cvpr}              % To produce the CAMERA-READY version
\definecolor{cvprblue}{rgb}{0.21,0.49,0.74}
\usepackage[pagebackref,breaklinks,colorlinks,allcolors=cvprblue]{hyperref}
\usepackage{diagbox}

\usepackage{multirow}
\usepackage[table]{xcolor}
\usepackage{mwe}
\usepackage{xcolor}      % 颜色支持
\usepackage{booktabs}    % 专业表格线
\usepackage{colortbl}    % 表格着色
\usepackage{bm}          % 粗体数学符号
\usepackage{array}
\usepackage{booktabs}
\usepackage{tabularx}

\def\paperID{920} % *** Enter the Paper ID here
\def\confName{CVPR}
\def\confYear{2026}

\title{Boosting Reasoning in Large Multimodal Models via Activation Replay}

\author{Yun Xing$^1$ \ \ \ \ \ Xiaobin Hu$^{2,\dagger}$ \ \ \ \ \ Qingdong He$^3$ \ \ \ \ \ Jiangning Zhang$^4$\\Shuicheng Yan$^{2,\dagger}$ \ \ \ \ \ Shijian Lu$^{1,\dagger}$ \ \ \ \ \ Yu-Gang Jiang$^5$ \\
$^1$Nanyang Technological University \ \ \ \ \ $^2$National University of Singapore\\$^3$Tencent Youtu Lab \ \ \ \ \ $^4$Zhejiang University \ \ \ \ \ $^5$Fudan University\\
{\tt\small \{xing0047, shijian.lu\}@ntu.edu.sg \ \ \{ben0xiaobin0hu1, yansc\}@nus.edu.sg}
}

\newcommand{\purple}[1]{{\textcolor{black}{#1}}}

\definecolor{darkred}{RGB}{139,0,0}
\definecolor{darkgreen}{RGB}{85,139,47}
\definecolor{forestgreen}{RGB}{34,139,34}
\definecolor{mydeepgreen}{RGB}{0,100,0}
\newcommand{\tabup}[1]{$_{\color{mydeepgreen}{#1\uparrow}}$}
\newcommand{\tabdown}[1]{$_{\color{darkred}{#1\downarrow}}$}

\begin{document}
%%%%%%%%%%%%%%% TOREMOVE %%%%%%%%%%%%%%% 
% \maketitle
% \begin{center}
% \centerline{\url{https://evolvinglmms-lab.github.io/OpenMMReasoner/}}
% \end{center}
%%%%%%%%%%%%%%% TOREMOVE %%%%%%%%%%%%%%% 

\twocolumn[{%
   \renewcommand\twocolumn[1][]{#1}%
   \maketitle
   \vspace{-10mm}
   \begin{center}
    \centering
    \centerline{\url{https://github.com/latentcraft/replay.git}}
   \end{center}%
}]

\def\thefootnote{$\dagger$}\phantomsection\footnotetext{Corresponding author}\def\thefootnote{\arabic{footnote}}
\begin{abstract}
Recently, Reinforcement Learning with Verifiable Rewards (RLVR) has emerged as an effective approach to incentivizing reasoning capability in Large Multimodal Models (LMMs), while the underlying mechanisms behind this post-training paradigm are poorly understood. We begin by exploring how input activations are affected by RLVR through the perspective of logit lens. Our systematic investigations across multiple post-trained LMMs suggest that RLVR shifts low-entropy activations unexpectedly, while high-entropy ones are less affected. We further demonstrate that such phenomena are associated with LMM reasoning by controlled experiments, suggesting a potentially beneficial role of modulating low-entropy activations. To this end, we propose \textbf{Activation Replay}, a novel simple yet effective training-free approach that boosts multimodal reasoning of post-trained LMMs without requiring expensive policy optimization. Our design involves manipulation of visual tokens at test time, replaying low-entropy activations from the input context of base LMMs to regulating the RLVR counterparts. Activation Replay triggers better reasoning across diverse scenarios, including mathematics, o3-like visual agents, and video reasoning. We further show that Activation Replay boosts Pass@K and mitigates narrower reasoning coverage of RLVR. Our design is compared against alternative choices, such as replaying high-entropy activations instead of low-entropy ones, or direct cross-model intervention instead of manipulating input tokens, demonstrating the superiority of our implementation. Codes will be made publicly available. 
\end{abstract}
\vspace{-4mm}    
\section{Introduction}
\label{sec:intro}

% Large Multimodal Models (LMMs)~\citep{liu2023llava,dai2023instructblip,bai2025qwen2_5_vl,zhu2025internvl3,guo2025seed1d5vl,abdin2024phi4} have achieved inspiring advances in understanding multimodal interactions from users. Recently, Reinforcement Learning with Verifiable Rewards (RLVR) has been playing an effective role in incentivizing step-by-step reasoning capacity in LMMs through Group Relative Policy Optimization (GRPO)~\citep{shao2024deepseekmath}, a post-training policy gradient update algorithm proven effective for LLM reasoning. Existing studies~\citep{meng2025mmeureka,wang2025vlrethinker,chen2025revisualr1} have made sufficient contributions in providing effective training recipes to address various reasoning scenarios, such as math-related~\citep{lu2021geometry3k} solving, chart understanding~\citep{masry2022chartqa}, or long video comprehension~\citep{chen2025scaling}. 

% Large Multimodal Models (LMMs)~\citep{liu2023llava,dai2023instructblip,bai2025qwen2_5_vl,zhu2025internvl3,guo2025seed1d5vl,abdin2024phi4} 

\noindent {Large Multimodal Models (LMMs)~\cite{liu2023llava,dai2023instructblip,bai2025qwen2_5_vl,zhu2025internvl3,guo2025seed1d5vl,abdin2024phi4} have been playing crucial roles in understanding multimodal interactions from users, across various domains such as images, videos~\cite{damonlpsg2024videollama2}, or 3-D inputs~\cite{zhu2024llava3d}. Recently, Reinforcement Learning with Verifiable Rewards (RLVR)~\cite{shao2024deepseekmath} has been verified as an effective approach that incentivizes reasoning capability in LMMs~\cite{meng2025mmeureka,wang2025vlrethinker}, characterized by long chain-of-thought verifications before arriving at the final answers. This approach opens new frontiers for existing LMMs to tackle challenging multimodal scenarios, such as mathematics~\cite{wang2024mathvision,qiao2024wemath,xiao2024logicvista}, agentic~\cite{zheng2025deepeyes,shen2025zoomeye,lai2025minio3} or video reasoning~\cite{chen2025scaling,feng2025videor1,cheng2025videoholmes}.}

\begin{figure}[tp]
    \centering
    \vspace{-6mm}
    \includegraphics[width=1\linewidth]{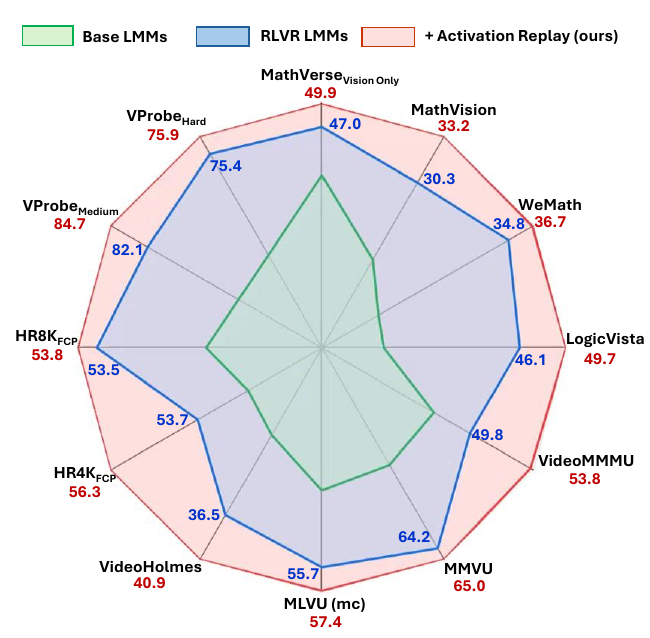}
    \vspace{-7mm}
    \caption{{\textbf{Lidar Plot for Performance Gains.} RLVR LMMs include VL-Rethinker, DeepEyes and Video-R1~\cite{wang2025vlrethinker,zheng2025deepeyes,feng2025videor1}. Our method boosts multimodal reasoning across diverse tasks consistently in training-free manner.}}
    \vspace{-2mm}
    \label{fig:lidar_gain}
\end{figure}

{Sufficient efforts have been made in effective post-training recipes~\cite{meng2025mmeureka,wang2025vlrethinker,chen2025revisualr1} across a wide spectrum of input domains. Early studies focus on enabling reasoning of LMMs via tailored chain-of-thought verification data and recipes with simple rule-based rewards~\cite{meng2025mmeureka}. More recent studies explore further on explicitly incentivizing reasoning patterns such as reflection~\cite{wang2025vlrethinker,wan2025srpo}, or optimizing training efficiency by addressing vanishing gradients~\cite{shen2025vlmr1} or supervising minority fork tokens~\cite{wang2025qwenforktoken}. These approaches have been solid explorations for incentivizing reasoning of LMMs with dedicated training.}

{In spite of broad explorations of RLVR recipes, it is still unclear on how the inner workings of LMMs are generally affected by such post-training, especially how multimodal contexts are modified unexpectedly. We investigate how input activations are shifted through the perspective of logit lens~\cite{nanda2021logitlens,neo2025towards}. The input activations of LMMs fundamentally determine how post-trained LMMs perform, especially for reasoning scenarios where decoding over input contexts involves long decoding steps.} 

\begin{figure}[t]
    \centering
    \vspace{-1mm}
    \includegraphics[width=\linewidth]{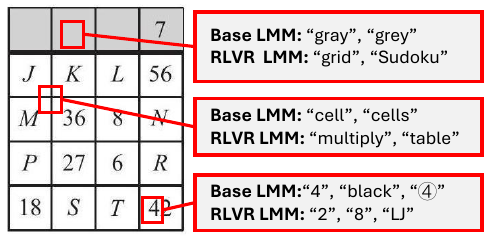}
    \vspace{-6mm}
    \caption{{\textbf{Qualitative Case on Logit Lens}. The math visual input is from~\cite{wang2024mathvision}. Top-2 or Top-3 predictions (words) of input activations shift from base to RLVR couterpart.}}
    \vspace{-2mm}
    \label{fig:logit_lens}
\end{figure}

{We first explore on how input activations are affected by RLVR, starting with observing the shifts of input activations. We discover evidences that top predictions through \textit{logit lens} are shifted after RLVR (Figure~\ref{fig:logit_lens}). Furthermore, we explore the effect of RLVR on Kullback-Leibler divergence, the metric that quantitatively measures token-level distributional shifts, and ablate the analysis by base LMM entropy percentiles~\cite{wang2025qwenforktoken}. To this end, we find high-entropy input activations are less affected by RLVR, as compared to low-entropy input activations (Figure~\ref{fig:rlvr_effect}). }

{We perform studies from two perspectives to investigate the distinctive role of low- and high-entropy input activations in multimodal reasoning. On the one hand, for the \textit{perturbation study}, we manually disrupt the inputs with random noises to synthesize variations on activations of post-trained LMMs, {so that KL divergence shifts over the same inputs vary}. The goal is to observe how the perplexity of sampled correct responses are influenced. Qualitative cases suggest that smaller KL divergence shifts on low-entropy activations leads to a lower perplexity of correct reasoning paths (Figure~\ref{fig:low_reason_effect}), while incorrect ones are vise versa (\purple{Appendix}). On the other, we perform a straightforward \textit{intervention study} to directly interrupt the forward pass of RLVR post-trained LMMs with low-entropy activations from their base LMMs, while keep the high-entropy ones. {Despite the representations of RLVR and base are largely different, we find that such strategy is beneficial in some cases}. Both studies lead to that regularization on RLVR low-entropy activations benefits multimodal reasoning.}

{On top of the abovementioned analysis, we propose Activation Replay, a simple yet effective training-free solution that boosts multimodal reasoning of post-trained LMMs, without requiring expensive policy optimization. By modulation of visual tokens at test time, Activation Replay enforces RLVR low-entropy activations to mimick the distributions from paired base LMMs at test time. We validate our approach across diverse LMMs post-trained by RLVR and diverse reasoning scenarios, including mathmatics~\cite{lu2023mathvista,qiao2024wemath,zhang2024mathverse,wang2024mathvision,xiao2024logicvista}, multi-turn agents that perform visual search in high-resolution images~\cite{wang2024hrbench}, and video reasoners that think across frames~\cite{feng2025videor1}, where our approach showcases consistent performance gains across these setups (Figure~\ref{fig:lidar_gain}). We further demonstrate that our design boosts Pass@K~\cite{brown2024passk}, mitigating narrower reasoning coverage of RLVR~\cite{yue2025limit-of-rlvr}. We summarize our contributions as follows,}

% \begin{figure}[!hbtp]
%     \centering
%     \includegraphics[width=1\linewidth,height=6cm]{example-image-a}
%     \caption{{An illustration of Logit Lens.}}
%     \label{fig:logit_lens}
% \end{figure}

% \begin{figure}[h]
%     \centering
%     \includegraphics[width=\textwidth,height=6cm]{example-image-a}
%     % \caption{Schematic view of how \textit{hidden entropy} is computed (b) demonstrative examples, where base model answers correctly and RLVR model answers falsely with high \textit{hidden entropy}. \blue{A training schedule, with eight checkpoints. For problems where base LMM reason correctly, RLVR fails, and it appears they are correlated somehow. Or can provide base LMM output and see how RLVR change the probability of this output.}}
%     % \caption{The negative effect of GRPO on the perplexity of responses predicted by VL-Rethinker-7B~\citep{wang2025vlrethinker}. The x-axis represents checkpoints from different training steps. As the training progresses, the LMM gradually increases their \textit{hidden entropy} of the demonstrated example inputs, while deviates from correct answers.}
%     \caption{\red{TODO} The negative effect of GRPO on the perplexity of responses. The x-axis represents checkpoints from different training steps. The results show that increased \textit{hidden entropy} of demonstrated example inputs is associated with deviation from correct answers during training. Further quantitative evidences are provided in Section~\ref{sec:2_2}.}
%     \label{fig:main1}
% \end{figure}

\begin{itemize}[leftmargin=1.5em, itemsep=0.3em]
    \item {We carry out systematic studies about how RLVR affects LMM input activations. We observe that low-entropy activations are affected more than high-entropy ones, and we reveal their associations with LMM reasoning behaviors through the perturbation and intervention studies.}
    \item {We propose Activation Replay, a novel training-free approach that involves manipulating visual tokens at test time, specifically, replaying low-entropy activations from the input context of base LMMs to regulate the RLVR counterparts.}
    \item {We test our approach over multiple multimodal reasoning scenarios, demonstrating that Activation Replay is effective across mathematical, o3-like agentic, and video reasoning. We additionally show that our design boosts Pass@K, mitigating narrower reasoning coverage of RLVR.}
\end{itemize}

\section{Motivation}
\label{sec:motivation}

{In this section, we discuss how LMM input activations are shifted by RLVR unexpectedly, from the perspective of \textit{logit lens}~\cite{nanda2021logitlens,neo2025towards}, and how these shifts are associated with LMM reasoning. We first briefly introduce how \textit{logit lens} is defined and then provide qualitative and quantitative comparisons (Figure~\ref{fig:logit_lens} and~\ref{fig:rlvr_effect}) on the shifts after RLVR. For exploring how reasoning is affected, we design perturbation and intervention studies, for which details are included in this section, followed by conclusion of our findings.}  

\subsection{Background}

\noindent {\textbf{Notation}. Consider an LMM $\mathcal{F}$, which typically comprises a visual encoder $\mathcal{V}$, a multimodal connector $\mathcal{F}_c$, and an LLM $\mathcal{M}$ with $L$ layers. The input to an LMM is a multimodal sequence of visual tokens $\{v_1, v_2, ..., v_n\}$ and text tokens from user instructions $\{t_1, t_2, ..., t_m\}$. During decoding, the LMM $\mathcal{F}$ processes the concatenated multimodal sequence $\{v_1, v_2, ..., v_n, t_1, t_2, ..., t_m\}$, namely a multimodal context~\cite{mei2025surveycontextengineering}, followed by autoregressively generated text tokens $\{y_1, y_2, ..., y_{k-1}\}$, to predict the next token $y_k$. At layer $l$, the visual input activations over layers are noted as $h_{l,i}$ (where $h_{0,i}$ corresponds to input tokens since it refers to the first layer). For the sake of clarity, we differentiate the input activations from base LMM and its RLVR counterpart with $h_{l,i}^b$ and $h_{l,i}^r$, respectively.}

\noindent {\textbf{Logit Lens}. Following previous studies~\cite{nanda2021logitlens,neo2025towards} we use norm $\mathcal{N}$ and vocabulary head $\mathcal{V}$ to project activations to the LMM output space. As in Figure~\ref{fig:logit_lens}, \textit{logit lens} interprets LMM activations when processing inputs. By applying the same mappings to activations from intermediate layers as that to the last layer, the technique reveals their distributions over LMM vocabulary and how such distributions converges over layers to the final prediction.}
\begin{equation}
    p_{l,i} = \operatorname{softmax}(\mathcal{V}(\mathcal{N}(h_{l,i})))
\end{equation}

\noindent {where $h_{l,i}$ denotes i-th activation at layer $l$.} 

\noindent {\textbf{Entropy}}. The perspective naturally motivates the exploration of activation entropy~\cite{farquhar2024nature}, which serves as a quantitative proxy for uncertainty of activations.  Existing studies have demonstrated the role of high-entropy minority tokens diverging reasoning paths~\cite{wang2025qwenforktoken}. Our study in input activations compensates for their findings, especially clarifying the roles of low-entropy activations. The entropy of an activation at layer $l$ and position $i$ is defined by,
\begin{equation}
    \label{eq:entropy}
    e_{l, i} = -p_{l,i} \cdot log(p_{l,i})
\end{equation}

\subsection{RLVR on Input Activations}
\label{sec:2_2}

\noindent {In this section, we provide a comparison over input activations from base LMMs and their paired RLVR post-trained LMMs both qualitatively and quantitatively.}

\noindent {\textbf{Top Predictions Shift}. To quantify the effects of RLVR on visual contexts, we first measure how the top predictions change from base and RLVR post-trained LMMs with exactly the same inputs, where the top prediction refers to the token sampled with the highest probability of distributions. A qualitative comparison is provided in Figure~\ref{fig:logit_lens}. It is clear that the top predictions are shifted after the RLVR. suggesting unexpected changes of RLVR over input context, which is fundamental for correct reasoning paths of LMMs. 
\begin{figure}[!ht]
    \centering
    \vspace{-1mm}
    \includegraphics[width=0.88\linewidth]{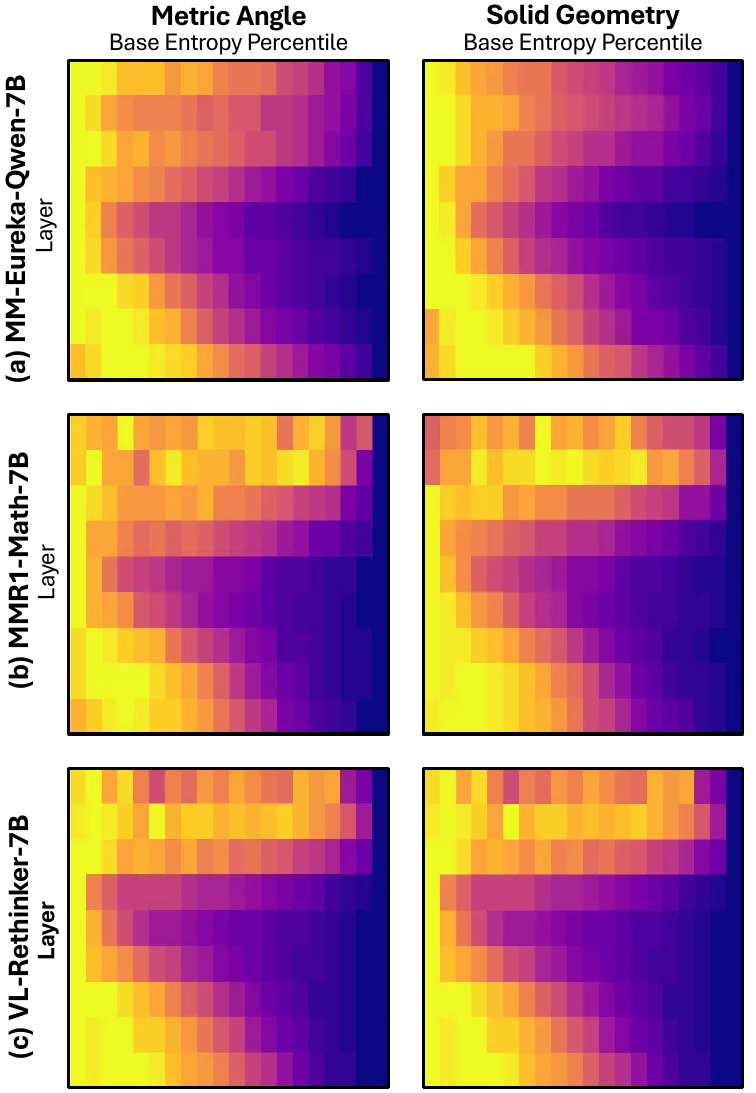}
    \vspace{-2mm}
    \caption{{\textbf{How LMM Input Activations are Affected after RLVR}. From left to right in subplots are \textit{low} to \textit{high} base LMM entropy. The shifts of KL divergence is normalized layerwise for illustration purpose. Brighter color suggests relatively more severe shifts on KL divergence. }}
    \vspace{-2mm}
    \label{fig:rlvr_effect}
\end{figure}

\noindent \textbf{KL Divergence}. Aside from top predictions by \textit{logit lens}, we further investigate the shifts over token-level KL divergence, as defined in Equation~\ref{eq:kl} for base and RLVR post-trained LMMs. Meanwhile, motivated by~\cite{wang2025qwenforktoken}, we explore how low- and high-entropy activations are affected differently, considering their different roles. Quantitative comparisons of three reasoning LMMs over two types of data are presented in Figure~\ref{fig:rlvr_effect}. We point out that the KL divergence is more informative than top predictions, since consistent top predictions between a pair of activations do not guarantee low KL divergence. Our findings suggest that across multiple reasoning LMMs and input domains, low-entropy activations are relatively shifted more than high-entropy activations over KL divergence. 
\begin{equation}
\label{eq:kl}
D_{kl}(P_{base} \| P_{rlvr}) = \sum p_{l,i}^b \log \frac{p_{l,i}^b}{p_{l,i}^r}
\end{equation}

\subsection{Perturbation Study}
\begin{figure*}[!htbp]
    \centering
    \vspace{-8mm}
    \includegraphics[width=\linewidth]{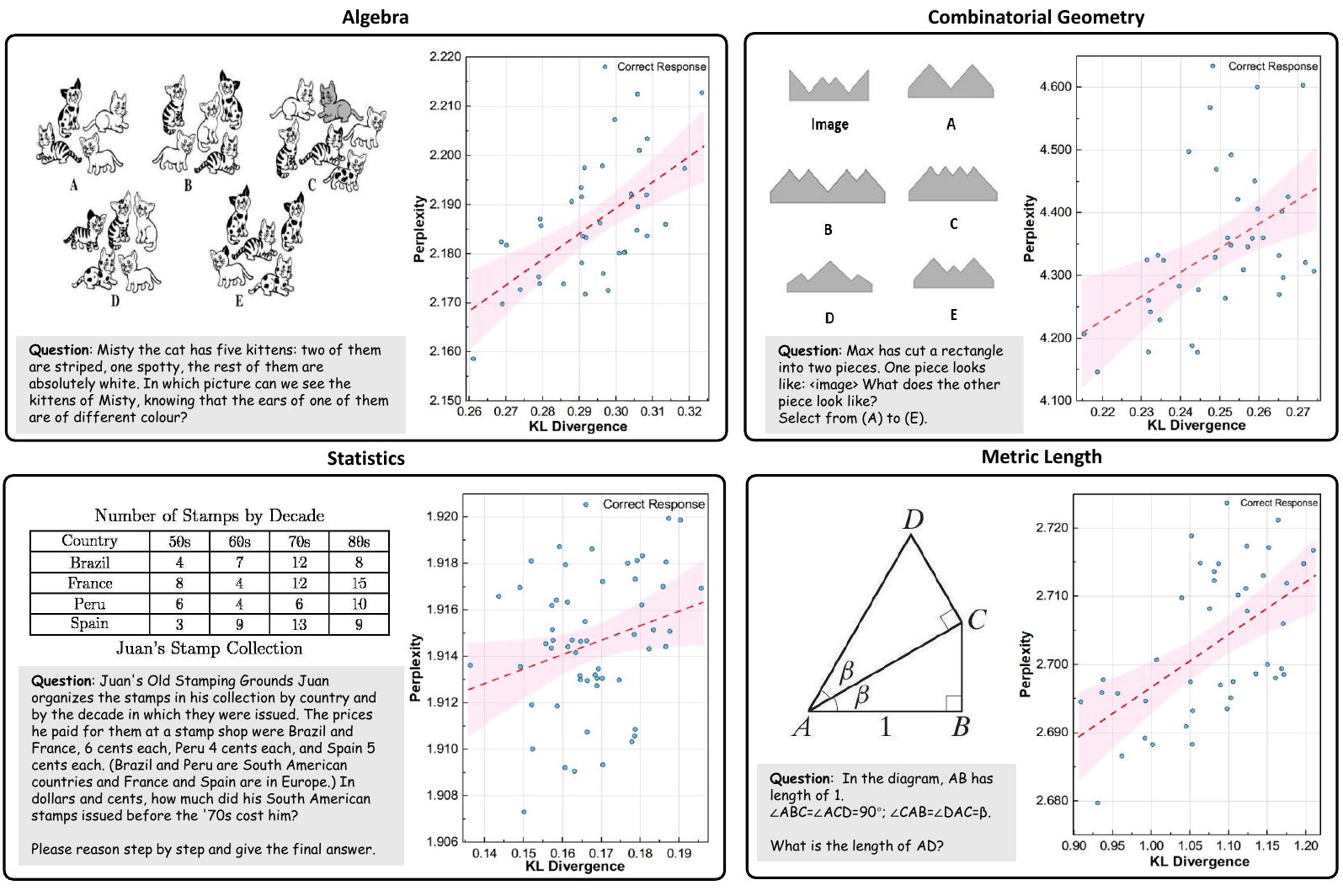}
    \vspace{-7mm}
    \caption{{\textbf{Perturbation Study}. We explore how low-entropy activations affects multimodal reasoning.}}
    \vspace{-2mm}
    \label{fig:low_reason_effect}
\end{figure*}

To reveal how KL divergence shifts in input activations are associated with LMM reasoning, we perform a simple case study. Specifically, given a multimodal input, we forward it through a post-trained LMM~\cite{meng2025mmeureka} and sample multiple responses, including correct and incorrect ones. We synthesize variations over input activations by interrupting inputs with random noises. We measure perplexity from reasoning LMNs over this response, while lower perplexity indicates higher probability. Four cases from different math domains are given in Figure~\ref{fig:low_reason_effect}. These cases suggest that when the KL divergence of low-entropy activations shifts less drastically from the base (left in subplots of Figure~\ref{fig:low_reason_effect}), the perplexity of the correct responses decreases and that of the incorrect responses increases, encouraging the LMMs to correct output. Analyses on incorrect responses are given \purple{in Appendix}.

\subsection{Intervention Study}

\noindent {To further dissect the role of low- and high-entropy activations from base LMMs, we perform a striaghtforward cross-model intervention study, forcing RLVR post-trained LMMs to reason over a combination of base and RLVR activations. We try two combinations, low-entropy activations from base and high from RLVR; high-entropy activations from base and low- from RLVR. The quantitative results of four LMMs are presented in Table~\ref{tab:tf_intervention_study}.} 

\begin{table}[!ht]
    \centering
    \caption{\textbf{Intervention Study}. We provide quantitative results of MM-Eureka and VL-Rethinker~\cite{meng2025mmeureka,wang2025vlrethinker} on MathVerse, MathVision, and WeMath~\cite{zhang2024mathverse,wang2024mathvision,qiao2024wemath}, shorted as ME, MN and WM, respectively. For strategy, \textit{Low} refers to inject low-entropy activations from base LMMs to RLVRs, while \textit{High} refers to the injection of high-entropy activations instead.}
    \label{tab:tf_intervention_study}
    \begin{small}
    \renewcommand{\arraystretch}{1.0}
        \begin{tabularx}{\linewidth}{p{0.2\linewidth}*{1}{>{\raggedright\arraybackslash}X}*{3}{>{\raggedright\arraybackslash}X}}
        \toprule
        \textbf{Model} & \textbf{Strategy} & \textbf{ME} & \textbf{MN} & \textbf{WM} \\
        \midrule
        \multirow{4}{*}{MM-Eureka} & Base & 41.1 & 25.5 & 24.5 \\
        & RLVR & 45.1 & 30.6 & 36.8 \\
        & Low & 45.1~\tabup{0.0} & 31.6~\tabup{1.0} & 36.6~\tabdown{0.2} \\
        & High & 42.1~\tabdown{3.0} & 27.6~\tabdown{4.0} & 32.8~\tabdown{4.0} \\
        % \midrule
        % \multirow{4}{*}{MMR1-Math} & \textit{base} & 41.1 & 25.5 & 24.5 \\
        % & \textit{rlvr} & 41.1 & 32.6 & 40.7 \\
        % & \textit{low} & 44.2 & 33.9 & 43.7 \\
        % & \textit{high} \\
        \midrule
        \multirow{4}{*}{VL-Rethinker} & Base & 41.1 & 25.5 & 24.5 \\
        & RLVR & 47.0 & 30.3 & 34.8 \\
        & Low & 47.5~\tabup{0.5} & 33.5~\tabup{3.2} & 35.3~\tabup{0.5} \\
        & High & 44.4~\tabdown{2.6} & 29.7~\tabdown{0.6} & 34.3~\tabdown{0.5} \\
        \bottomrule
        \end{tabularx}
    \end{small}
    \vspace{-2mm}
\end{table}

\noindent \textbf{Findings}. {Intriguingly, we observe that the intervening RLVR forward pass with low-entropy activations from the base LMMs improves reasoning performance, while the other setup, the intervening RLVR forward pass with high-entropy activations from the base LMMs downgrades performance. The phenomenon with high-entropy activations aligns with the recent study that high-entropy minorities encourage explorations~\cite{wang2025qwenforktoken}, while replacing them with those from base LMMs hurts reasoning. However, it is interesting to see on the input-level, that low-entropy activations play a role, which achieves better reasoning when co-works with high-entropy activations from reasoning LMMs}.

\noindent {\textbf{Limitations of direct intervention as an approach}. Despite it is quite effective to reasoning over a combination of activations from base and RLVR LMMs, the activations from base LMMs do not align the representations of RLVR post-trained LMMs, hindering its gains as a training-free solution (\purple{more evidences in Appendix}).}

\begin{figure*}[!ht]
    \centering
    \vspace{-7mm}
    \includegraphics[width=1\linewidth]{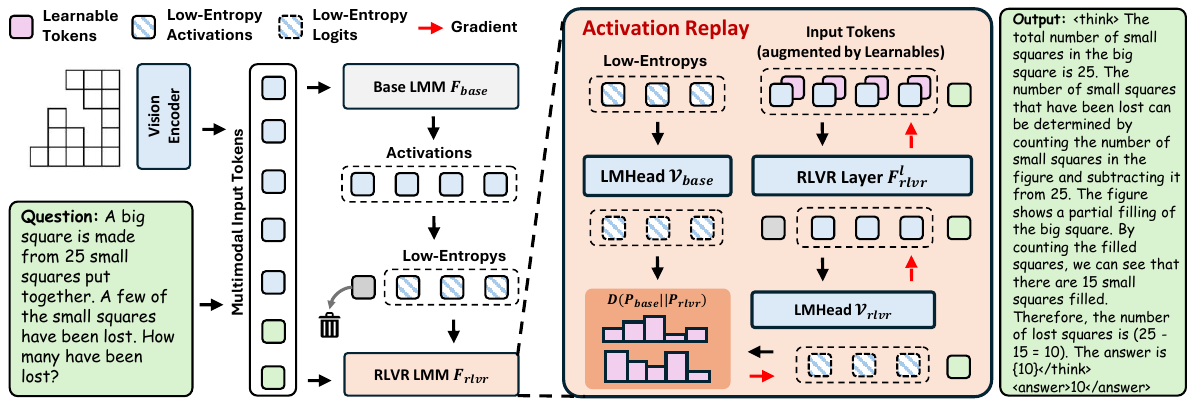}
    \vspace{-6mm}
    \caption{{\textbf{Overview of Activation Replay}. Activation Replay starts with feeding the multimodal inputs to base LMMs and obtain low-entropy input activations. For inputs to the RLVR LMM, our approach first adds zero-initialized learnable tokens to visual tokens. Then we manipulate these learnable tokens to minimize the token-level KL divergence between low-entropy activations from base LMMs and those from RLVR post-trained counterparts. }}
    \label{fig:approach}
    % \vspace{-4mm}
\end{figure*}
\section{Activation Replay}
\label{sec:activation_replay}

\noindent {On top of our observations, we propose a simple yet effective test-time approach that boosts multimodal reasoning in LMMs. Our core idea is to indirectly enforce RLVR input activations to mimick the counterparts from base LMMs. We achieve this by optimizing a set of learnable tokens based on a KL divergence term. } 

\noindent {\textbf{Learnable Tokens}. To minimize KL divergence of low-entropy input activations from RLVR at test time, we introduce a series of learnable tokens $\{x_1, x_2, ..., x_n\}$} that are zero-initialized. They augment multimodal input tokens by token-level addition. More specifically,
\begin{equation}
    \label{eq:apply}
    \hat{h}_{0,i}^r = h_{0,i}^r + x_i, i  \in [1, n]
\end{equation}

\noindent {\textbf{Low-entropy Threshold}. For input activations, we need to define which activations are low-entropy and which are high. Note that alternative choices are available when deciding low- against high-entropy. One straightforward strategy to determine this threshold is through a held-out validation set. Another solution is more dynamic, determining low-entropy activations with a fixed threshold $\tau$ and the maximum entropy of base LMM activations. We empirically find that both solutions perform well (\purple{comparison is in Appendix}). For the ease of simplicity, we resort to dynamic thresholding in our paper. The selection of low-entropy activations is defined as,}

\begin{equation}
    M_{l,i} = \begin{cases} 1, e_{l,i}^b < max(h_l^b) * \tau \\0, else
    \end{cases}
\end{equation}

\noindent where $e_{l,i}^b$ is the entropy of base LMM activation at layer $l$ and token $i$, defined in Equation~\ref{eq:entropy}. 

\noindent {\textbf{Replay Function}. As mentioned in Section~\ref{sec:2_2}, we project activations to the LMM output space by \textit{logit lens}, which involves a vocabulary head $\mathcal{F}_v$ and a normalization layer $\mathcal{F}_n$ (ignored in Figure~\ref{fig:approach} for simplicity). As illustrated in Figure~\ref{fig:approach}, the low-entropy activations from base LMM $\mathcal{B}$ are provided as target distributions, indirectly guided by \textit{logit lens} $\mathcal{F}_v^b$. The counterpart activations in RLVR are enforced to match the distribution with manipulating the learnable latent $x_i$. Specifically, }

\begin{equation}
\boldsymbol{x}_i \leftarrow \boldsymbol{x}_i- \alpha \nabla_{\boldsymbol{x}_i} (D_{kl}(P_{base} \| P_{rlvr}) \cdot M_{i}),
\end{equation}

\noindent {where $\alpha$ is the strength that controlling the guidance from LMM $\mathcal{B}$. By optimizing $\boldsymbol{l}_t$, we shift the distribution of input activations towards that of base LMMs.}

\noindent \textbf{Sample on Manipulated Input Activations}. After manipulations over visual tokens, we obtain a series of vector $x_i$, which we add to inputs as in Equation~\ref{eq:apply} during test time before any decoding, including greedy decoding when evaluating Pass@1 or sampling when Pass@K is measured.
\section{Experiments}

We first briefly describe evalution settings, then introduce the results of our approach over three multimodal reasoning scenarios in Section~\ref{sec:main_results}. In Section~\ref{sec:passk}, We then present the experiments of our approach on Pass@K~\cite{brown2024passk}, followed by ablations over two parameters in our training-free design in Section~\ref{sec:ablation}. For other experimental results, we refer readers to Appendix, such as clarification of extra inference time costs or case studies of o3-like agentic and video reasoning.
\begin{table*}[!ht]
    \centering
    \caption{Applying training-free \textit{Activation Replay} to existing reasoning LMMs. From left to right, MathVerse (Vision Only)~\cite{zhang2024mathverse}, MathVision~\cite{wang2024mathvision}, MathVista~\cite{lu2023mathvista}, DynaMath~\cite{zou2024dynamath}, WeMath~\cite{qiao2024wemath}, LogicVista~\cite{xiao2024logicvista}, MMMU~\cite{yue2024mmmu}, MMMUPro~\cite{yue2024mmmu}.}
    \vspace{-1mm}
    \label{tab:tf_main_math_8cols}
    \begin{small}
    \setlength{\tabcolsep}{4pt}
    \renewcommand{\arraystretch}{1.0}
        \begin{tabularx}{\linewidth}{>{\raggedright\arraybackslash}p{0.26\linewidth}*{8}{>{\raggedright\arraybackslash}X}}
        \toprule
        \textbf{Model} & \textbf{ME} & \textbf{MN} & \textbf{MA} & \textbf{DM} & \textbf{WM} & \textbf{LV} & \textbf{MU} & \textbf{MP}\\
        % \cmidrule(lr){2-2} \cmidrule(lr){3-3} \cmidrule(lr){5-5} \cmidrule(lr){6-6} \cmidrule(lr){7-7} \cmidrule(lr){8-8} \cmidrule(lr){9-9} \cmidrule(lr){10-10} \cmidrule(lr){11-11} & \textit{vision} & \textit{full} & \textit{mini} & \textit{mini} & \textit{worst} & \textit{strict} & \textit{full} & \textit{val} & \textit{full} & \textit{full} \\
        \midrule
        \multicolumn{9}{l}{\textit{Close-Source LMMs}} \\
        \midrule
        Gemini-2.0-Flash~\cite{team2023gemini} & 43.6 & 47.8  & 70.4 & % 42.1 
        - & 47.4 & 52.3 & 70.7 & 51.7  \\
        GPT-4o~\cite{hurst2024gpt4o} & 40.6 & 31.1 & 59.9 & % 34.5 
        63.7 & 42.9 & 64.4 & 51.9 & 69.1 \\
        Claude-3.5 Sonnet~\cite{claude3.5sonnet2024} & 52.0 & 41.3  & 59.8 & % 39.7 
        64.8 & 58.2 & 49.3 & 51.5 & 68.3 \\
        GPT-4o-mini~\cite{hurst2024gpt4o} & 30.0 & 27.3  & 55.1 & % 31.6 
        - & 31.4 & 41.4 & - & - \\
        \midrule
        \multicolumn{9}{l}{\textit{Open-Source LMMs}} \\
        \midrule
        
        % === Baseline Models ===
        Qwen2.5-VL-3B~\cite{bai2025qwen2_5_vl} & 17.5 & 16.1 & 48.0 & 42.7 & 10.8 & 26.6 & 53.1 & 31.6 \\
        LLaVA-OV-7B~\cite{li2024llavaov} & 17.6 & 17.0  & 62.6 & 34.3 & 17.7 & 32.0 & 48.8 & 24.1 \\
        Kimi-VL-16B~\cite{kimivl} & 34.1 & 21.8  & 66.0 & - & 32.3 & 42.7 & 57.0 & 46.3 \\
        InternVL3-8B~\cite{zhu2025internvl3} & 33.9 & 28.6  & 70.5 & 46.5 & 37.5 & 43.6 & 62.7 & - \\
        Qwen2.5-VL-7B~\cite{bai2025qwen2_5_vl} & 41.1 & 25.5  & 69.1 & 53.2 & 24.5 & 35.6 & 58.6 & 38.3 \\
        % Qwen2.5-VL-32B &&&&&&&& 70.0 & 49.5 \\
        QvQ-72B-Preview~\cite{qwen2024qvq} & 48.2 & 34.9  & 66.0 & - & 32.3 & 42.7 & 70.3 & - \\
        \midrule

        % === VLAA-Thinker-3B Group ===
        % VLAA-Thinker-3B & \gray{36.4} & \gray{24.4} & & \gray{61.0} & \gray{18.2} & \gray{33.8} & \gray{38.5} \\
        % & 36.2 & 23.8 & 24.0 & 59.9 & 11.6 & 30.7 & 39.1 & 51.8 & 32.7 & 24.2 \\
        % \rowcolor{gray!15}\textit{+ replay} & & & & & & & & & & \\
        % % & & 47.7 & 26.4 & 25.6  & 68.1 & 23.5 & 40.5 & 44.3 & 42.8 & 38.5 \\
        % % \rowcolor{gray!15}\textit{+ replay} \\
        % \midrule
        \multicolumn{9}{l}{\textit{RLVR Post-trained LMMs + Activation Replay}} \\
        \midrule
        % === MMR1-Math Group ===
        MMR1-Math-v0-7B~\cite{leng2025mmr1} & 41.1 & 32.6 & 71.0 & 53.8 & 40.7 & 43.6 & 56.7 & 41.3 \\
        \rowcolor{gray!15}\textit{+ replay} & 43.3~\tabup{2.2} & 33.9~\tabup{1.3} & 71.9~\tabup{0.9} & 54.2~\tabup{0.6} & 41.8~\tabup{1.1} & 44.5~\tabup{0.9} & 58.7~\tabup{2.0} & 41.3~\tabup{0.0} \\
        \midrule
        
        % === MM-Eureka Group ===
        MM-Eureka-Qwen-7B~\cite{meng2025mmeureka} & 45.1 & 30.6 & 73.0 & 51.8 & 36.8 & 49.2 & 58.7 & 39.5 \\
        \rowcolor{gray!15}\textit{+ replay} & 47.7~\tabup{2.6} & 31.5~\tabup{0.9} & 73.5~\tabup{0.5} & 52.4~\tabup{0.8} & 38.0~\tabup{1.2} & 51.0~\tabup{1.8} & 62.0~\tabup{3.3} & 40.0~\tabup{0.5} \\
        \midrule
        
        % === VL-Rethinker-7B Group ===
        VL-Rethinker-7B~\cite{wang2025vlrethinker} & 47.0 & 30.3 & 72.0 & 54.7 & 34.8 & 46.1 & 58.7 & 41.7 \\
        \rowcolor{gray!15}\textit{+ replay} & 49.2~\tabup{2.2} & 33.2~\tabup{2.9} & 72.4~\tabup{0.4} & 54.9~\tabup{0.2} & 36.7~\tabup{1.9} & 49.7~\tabup{3.8} & 60.0~\tabup{1.3} & 42.7~\tabup{1.0} \\
        \midrule

        % === VLAA-Thinker-7B Group ===
        % VLAA-Thinker-7B & 47.6 & 25.6 & 68.0 & 22.4 & 41.5 & 44.3 \\
        % \rowcolor{gray!15}\textit{+ replay} & 48.9 & 29.3 & 69.4 & & 40.1 & 46.1 &&&\\
        % \midrule

        % === WeThink-7B Group ===
        % WeThink-7B & \gray{44.7} & \gray{27.2} & & \gray{70.9} & \gray{24.4} & \gray{48.0} & \gray{53.0} \\
        % & 45.8 & & 24.0 & 70.2 & 25.0 & 44.5 & 50.3 \\
        % \rowcolor{gray!15}\textit{+ replay} & \textminus & \textminus & \textminus & \textminus & \textminus && &&\\
        % & 46.0 & 25.9 & 24.0 & 70.0 & 25.0 & 44.1 & 50.1 & 59.1 \\
        % \rowcolor{gray!15}\textit{+ replay} & \textminus & \textminus & \textminus & \textminus & \textminus && &&\\
        % \midrule

        % === Revisual-R1 Group ===
        % Revisual-R1-7B & \gray{53.6} & \gray{48.8} & & \gray{73.1} & \gray{27.5} & \gray{42.0} & \gray{52.3} \\
        %  & 52.3 & \textminus & 39.5 & \textminus  & \textminus &&&&\\
        % \rowcolor{gray!15}\textit{+ replay} & \textminus & \textminus & \textminus & \textminus & \textminus &&&&\\
        % \midrule

        % === OVR-7B Group ===
        % OVR-7B & 54.6 & 51.8 & 72.1 & 33.5 & 44.6 & 54.8 \\
        % \textit{+ replay} & \textminus & \textminus & \textminus & \textminus & \textminus & \textminus \\
        % \midrule
        
        % === Larger Models ===
        MM-Eureka-Qwen-32B~\cite{meng2025mmeureka} & 50.5 & 35.2 & 72.1 & 62.1 & 36.9 & 52.8 & 59.3 & 49.6 \\
        \rowcolor{gray!15}\textit{+ replay} & 52.4~\tabup{1.9} & 35.5~\tabup{0.3} & 74.0~\tabup{1.9} & 61.8~\tabdown{0.3} & 37.6~\tabup{0.7} & 54.6~\tabup{1.8} & 63.2~\tabup{3.9} & 51.0~\tabup{1.4} \\
        
        % \midrule
        % VL-Rethinker-72B & \gray{63.5} & & \gray{44.9} & & \gray{80.4} & \textminus & \textminus & \textminus \\
        % \rowcolor{gray!15}\textit{+ replay} & \textminus & \textminus & \textminus & \textminus & \textminus & \textminus &&&& \\
        \bottomrule
        \end{tabularx}

        \vspace{-1mm}
    \end{small}
\end{table*}
\begin{table}[!htbp]
    \centering
    % \vspace{-4mm}
    \caption{Agentic LMMs that think with images. We test effectiveness of our approach on the recent agentic LMM~\cite{zheng2025deepeyes} and on HRBench (FCP)~\cite{wang2024hrbench}, and VisualProbe (Medium and Hard)~\cite{lai2025minio3}. }
    \label{tab:tf_main_deepeyes}
    \begin{small}
    \renewcommand{\arraystretch}{1.0}
        \begin{tabularx}{\linewidth}{p{0.32\linewidth}*{4}{>{\raggedright\arraybackslash}X}}
        \toprule
        {\textbf{Model}} & 
        \textbf{H4} & \textbf{H8} & \textbf{VM} & \textbf{VH} \\
        \midrule
        GPT-4o~\cite{hurst2024gpt4o}  & 48.0 & 49.0 & - & - \\
        ZoomEye~\cite{shen2025zoomeye} & 55.0 & 50.0 & - & - \\
        \midrule
        LLaVA-OV-7B~\cite{li2024llavaov} & 54.0 & 52.3 & - & - \\
        \midrule
        Qwen2.5-VL-7B~\cite{bai2025qwen2_5_vl} & 52.2 & 51.8 & 26.0 & 23.9 \\
        DeepEyes-7B~\cite{zheng2025deepeyes} & 53.7 & 53.5 & 29.8 & 35.1 \\
        \rowcolor{gray!15}\textit{+ replay} & 56.3~\tabup{2.6} & 53.8~\tabup{0.3} & 32.8~\tabup{3.0} & 35.9~\tabup{0.8} \\
        \bottomrule
        \end{tabularx}
    \end{small}
\end{table}
\begin{table}[!htbp]
    \centering
    \caption{Apply Activation Replay to 16-frame video reasoning LMMs. From left to right, we evaluate our approach on VideoMMMU, MLVU, MMVU (mc), VideoHolmes~\cite{hu2025videommmu,zhou2024mlvu,zhao2025mmvu,cheng2025videoholmes}.}
    \label{tab:tf_main_video_r1_4cols}
    \begin{small}
    \renewcommand{\arraystretch}{1.0}
        \begin{tabularx}{\linewidth}{>{\raggedright\arraybackslash}p{0.35\linewidth}*{4}{>{\raggedright\arraybackslash}X}}
        \toprule
        \textbf{Model} & \textbf{VU} & \textbf{ML} & \textbf{MV} & \textbf{VH} \\
        \midrule
        GPT-4o~\cite{hurst2024gpt4o} & 61.2 & 54.2 & 75.4 & 42.0 \\
        Claude-3.5~\cite{claude3.5sonnet2024} & 65.8 & - & - & 41.0 \\
        \midrule
        LLaVA-OV-7B~\cite{li2024llavaov} & 33.9 & - & 49.2 & -\\
        % \midrule
        % GPT-4o & -- & -- & -- & -- & -- & -- \\
        % Gemini-1.5 & -- & -- & -- & -- & -- & -- \\
        % \midrule
        % LLaVA-OV-7B~\cite{li2024llavaov} & 56.4 & 33.9 & 64.7 & 58.2 & 32.4 & -- \\
        % Qwen2-VL-7B~\cite{bai2025qwen2_5_vl} & -- & -- & -- & -- & -- & -- \\
        \midrule
        Qwen2.5-VL-7B~\cite{bai2025qwen2_5_vl} & 47.4 & 48.4 & 50.1 & 27.8 \\
        Video-R1-7B~\cite{feng2025videor1} & 49.8 & 55.7 & 64.2 & 36.5 \\
        \rowcolor{gray!15}\textit{+ replay} & 53.8~\tabup{4.0} & 57.4~\tabup{1.7} & 65.0~\tabup{0.8} & 40.9~\tabup{4.4} \\
        \bottomrule
        \end{tabularx}
    \end{small}
\end{table}
\subsection{Implementation Details}

{\textbf{Models}. To verify the effectiveness of our approach, we apply \textit{Activation Replay} over multiple recent reasoning-enhanced LMMs, including MMR1-Math, MM-Eureka, VL-Rethinker~\cite{leng2025mmr1,wang2025vlrethinker,meng2025mmeureka}. For MM-Eureka, we test our approach on two model scales, 7B and 32B~\cite{meng2025mmeureka}. For o3-like agentic reasoning, we use DeepEyes~\cite{zheng2025deepeyes} as our baseline. For video reasoning, we use Video-R1~\cite{feng2025videor1}. We refer readers to Appendix for more details about implementations.}

% \textbf{Comparisons}. For \textit{Activation Replay}, we copy and paste low-entropy tokens from instruct to RLVR LMMs by default. We compare replay low-entropy tokens with either random or high-entropy tokens to further justify our design choices. \blue{We also compare the proposed replay strategy with calibration approach, which is directly performed over low-entropy tokens in RLVR models}.

\noindent \textbf{Evaluation}. Our evaluation comprises multiple dimensions of reasoning cases, including mathematics, such as MathVerse, MathVision, MathVista, DynaMath, WeMath, and LogicVista, MMMU, MMMUPro~\cite{zhang2024mathverse,wang2024mathvision,lu2023mathvista,zou2024dynamath,qiao2024wemath,xiao2024logicvista,yue2024mmmu,yue2025mmmupro}. For o3-like agentic reasoning characterized by multi-turn visual search, we compare on HR-Bench~\cite{wang2024hrbench}. For video reasoning that involves complex multi-frames, we test multiple benchmarks, including VideoMMMU, MLVU, MMVU, Video-Holmes~\cite{hu2025videommmu,zhou2024mlvu,zhao2025mmvu,cheng2025videoholmes}. Following previous studies we use GPT-4o-mini~\cite{hurst2024gpt4o} as judge, except for measuring Pass@K we use Qwen3-30B-A3B~\cite{yang2025qwen3}.

\subsection{Main Results}
\label{sec:main_results}

\noindent {\textbf{Image-Level Mathmetical and Knowledge Reasoning}. The performance comparison on math reasoning are provided in Table~\ref{tab:tf_main_math_8cols}. Activation Replay consistently improves mathematical reasoning over tested datasets. Beyond boosting reasoning for small parameter-scale reasoning LMMs, our design also boosts reasoning for comparably stronger LMMs that are post-trained with stronger base, such as MM-Eureka-32B~\cite{meng2025mmeureka}. While various mathematical data requires diverse reasoning aspects, such as algebra, geometry or logic~\cite{xiao2024logicvista}, Activation Replay performs better consistently. On knowledge-intensive evaluations such as MMMU~\cite{yue2024mmmu}, our approach is also effective}.

\noindent {\textbf{Agentic Reasoning}. Agentic Reasoning is characterized by multi-turn explorations of visual cues and excels in perception of high-resolution images. We use the recent representative work DeepEyes~\cite{zheng2025deepeyes} and test our approach on HRBench. The results are given in Table~\ref{tab:tf_main_deepeyes}. Despite very distinct from mathematical reasoning in reasoning formats, Activation Replay still brings performance gains over this setup, especially on cross-object setup which are more challenging. Case study is given in \purple{Appendix}.}
\begin{figure}[t]
    \centering
    \vspace{-2mm}
    \includegraphics[width=0.94\linewidth]{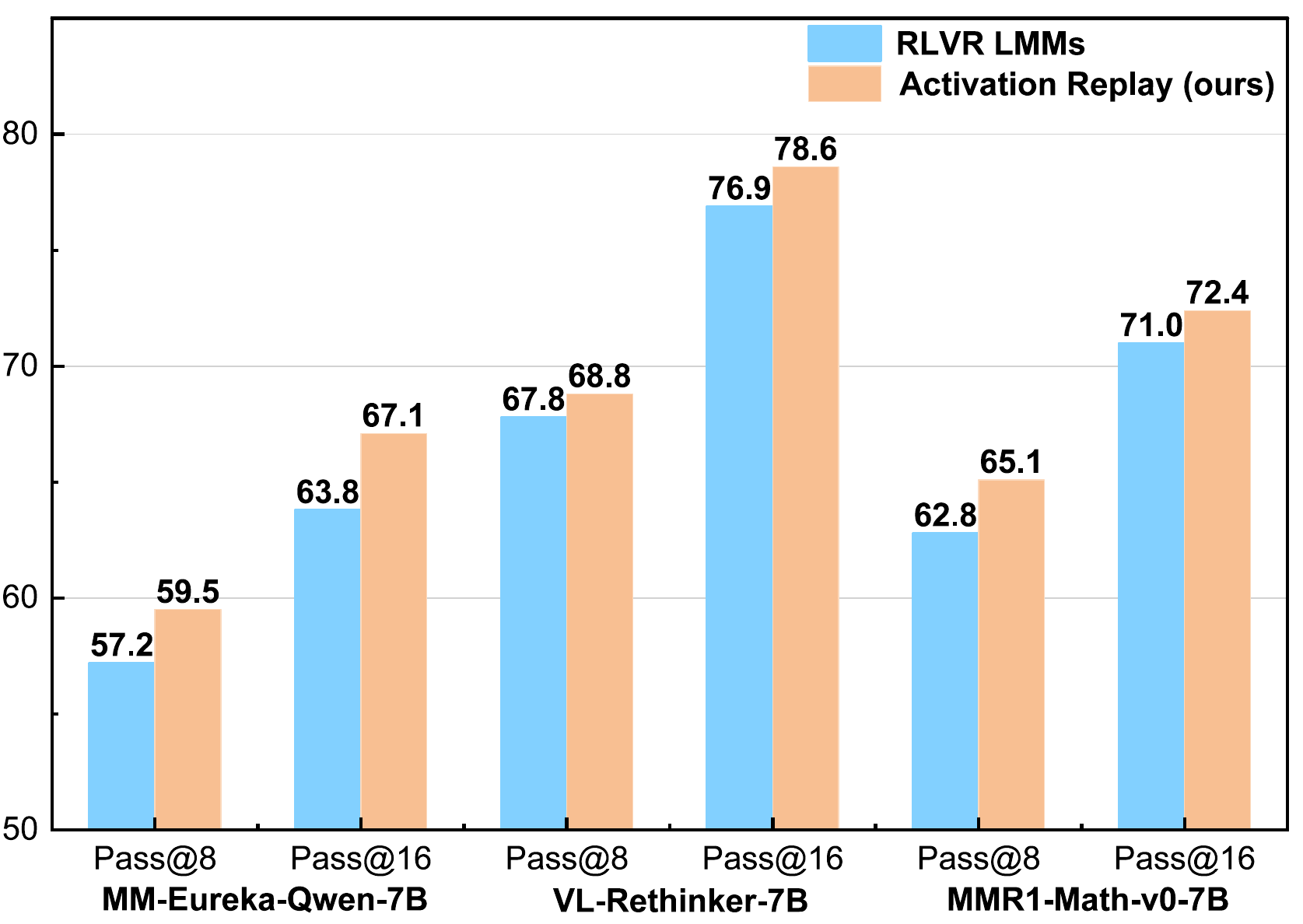}
    \vspace{-2mm}
    \caption{{The Effect of Activation Replay on Pass@K.}}
    \vspace{-2mm}
    \label{fig:tf_passk}
\end{figure}

\noindent \textbf{Video Reasoning}. Different from mathematical reasoning we evaluate on image-level, video reasoning requires cross-frame interaction. We use the recent Video-R1~\cite{feng2025videor1} and four datasets as the testbed. The results are given in Table~\ref{tab:tf_main_video_r1_4cols}. We observe consistent performance gains from multiple evaluation aspects, including basic understanding as MLVU, knowledge-intensive reasoning, and reasoning-intensive bench like VideoHolmes.

\subsection{Activation Replay Boosts Pass@K}
\label{sec:passk}
\begin{table}[!htbp]
    % \centering
    \vspace{-2mm}
    \caption{Ablation study with MathVerse (Vision Only)~\cite{zhang2024mathverse} on hyperparameters.}
    \label{tab:ablation_alpha_tau}
    \begin{small}
    \renewcommand{\arraystretch}{0.9}
        \begin{tabularx}{\linewidth}{p{0.25\linewidth}*{4}{>{\raggedright\arraybackslash}X}}
        \toprule
         % & & \multicolumn{3}{c}{} \\
        {Model} & $\gamma$ \textbackslash\ $\alpha$ & 10.0 & 20.0 & 40.0 \\
        \midrule
        \multirow{4}{*}{MMR1-Math} & 0.2 & 42.3~\tabup{1.2} & 42.5~\tabup{1.4} & 40.7~\tabdown{0.4} \\
        & 0.4 & \textbf{43.3}~\tabup{2.2} & 41.8~\tabup{0.7} & 42.1~\tabup{1.0} \\
        & 0.6 & 41.4~\tabup{0.3} & 41.4~\tabup{0.3} & 42.4~\tabup{1.3} \\
        & 0.8 & 41.5~\tabup{0.4} & 43.2~\tabup{2.1} & 42.0~\tabup{0.9} \\
        \midrule
        \multirow{4}{*}{MM-Eureka} & 0.2 & 45.4~\tabup{0.3} & 46.1~\tabup{1.0} & \textbf{47.7}~\tabup{2.6} \\
        & 0.4 & 47.1~\tabup{2.0} & \textbf{47.5}~\tabup{2.4} & 46.5~\tabup{1.4} \\
        & 0.6 & 46.2~\tabup{1.1} & 44.9~\tabdown{0.2} & 47.2~\tabup{2.1} \\
        & 0.8 & 46.1~\tabup{1.1} & 45.1~\tabup{0.0} & 47.0~\tabup{1.9} \\
        \midrule
        % \multicolumn{4}{c}{\textbf{VL-Rethinker}} \\
        % \midrule
        \multirow{4}{*}{VL-Rethinker} & 0.2 & 47.3~\tabup{0.3} & 48.1~\tabup{1.1} & \textbf{49.2}~\tabup{2.2} \\
        & 0.4 & 49.0~\tabup{2.0} & 47.8~\tabup{0.8} & 48.4~\tabup{1.4} \\
        & 0.6 & 49.1~\tabup{2.1} & 46.5~\tabdown{0.5} & 48.7~\tabup{1.7} \\
        & 0.8 & 47.1~\tabup{0.1} & 47.2~\tabup{0.2} & 47.2~\tabup{0.2} \\
        \bottomrule
        \end{tabularx}
    \end{small}
    \vspace{-2mm}
\end{table}

\noindent {\textbf{Metric}. A recent study~\cite{yue2025limit-of-rlvr} pinpoints that RLVR mainly improves sampling efficiency but does not elicit novel reasoning capability beyond base models. Unlike greedy decoding that reflect average case performance and practical utility, Pass@K tests the reasoning boundary of a specific model. Given a problem, we sample $k$ outputs from a model. The Pass@K for this problem is correct if at least one of the $K$ outputs is correct.

\noindent \textbf{Results}. We test four reasoning post-trained LMMs over two mathematical datasets, MathVision MINI and MathVista. Our results are presented in Figure~\ref{fig:tf_passk}. Consistently on four LMMs, our Activation Replay boosts Pass@K. Especially, on Pass@16, we observe large margins of 5.0. Without additional training, or architectural modifications, Activation Replay brings performance gains on Pass@K. \purple{More details are given in Appendix.}

\subsection{Ablations}
\label{sec:ablation}

\noindent {\textbf{Threshold $\tau$}. The ablation over thresholds investigates the sensitivity of Activation Replay to the proportion of activations selected for replay. As shown in Table~\ref{tab:ablation_alpha_tau}, we vary the threshold from 0.2 to 0.8, while higher thresholds progressively restrict the number of activations used. The consistent trend across datasets indicates that small thresholds keep the better performance gains while avoiding the redundancy introduced by replaying all activations. By contrast, excessively large thresholds (\textit{e.g.,} 0.8) with inclusion of replaying high-entropy activations more possibly lead to less satisfying results, suggesting that high-entropy activations can hinder the approach effectiveness. These findings highlight the role of low- and high-entropy activations when applying our approach.}

\noindent {\textbf{Strength $\alpha$}. We further examine how the strength of $alpha$ affect Activation Replay. From one perspective, we note that nearly all $alpha$ selection improves over the baseline. From another, we point out that Activation Replay benefits from larger strength $\alpha$, regardless of selection over threshold $\tau$, further justifying the effectiveness of our core idea.}

\subsection{Case Study}
\begin{figure*}[tp]
    \centering
    \vspace{-8mm}
    \includegraphics[width=1\linewidth]{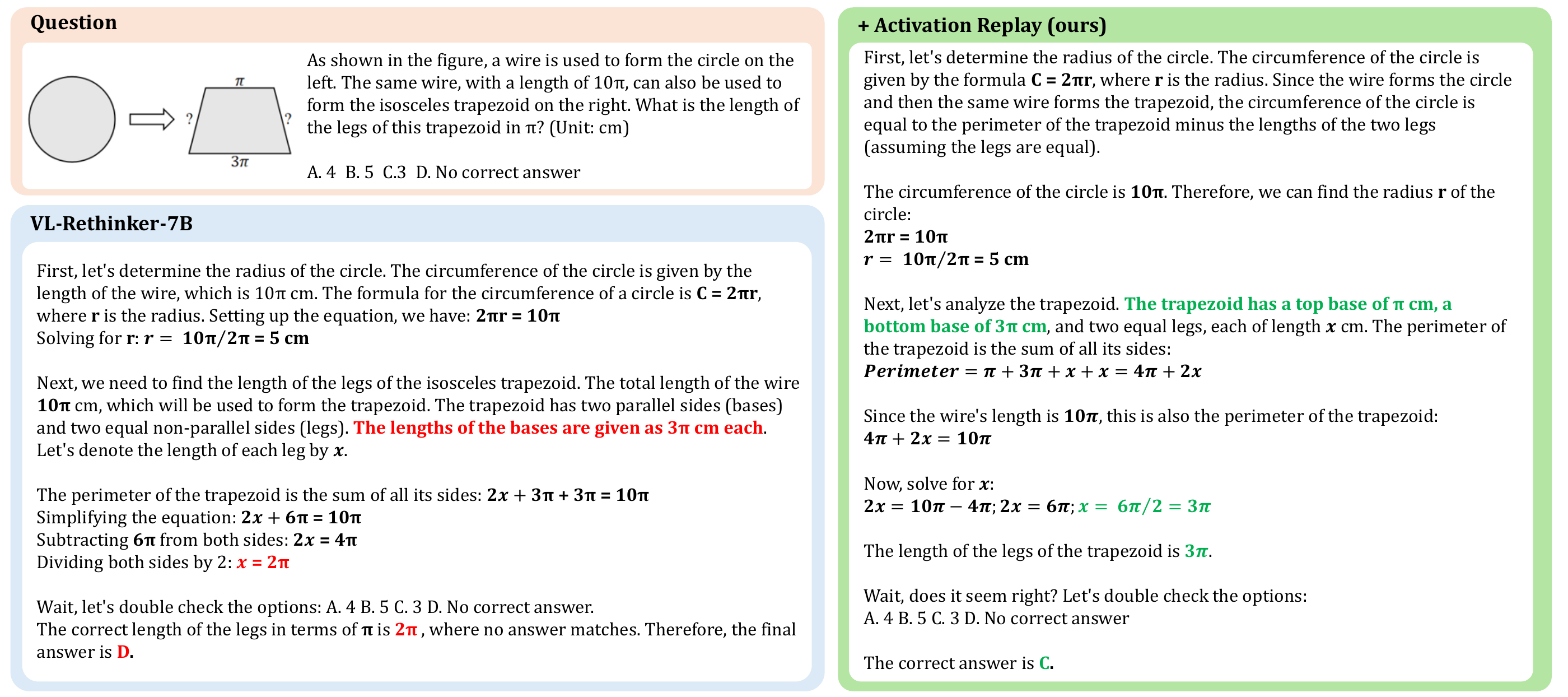}
    \vspace{-5mm}
    \caption{{Case Study. The RLVR LMM (VL-Rethinker-7B) fails to correctly reason in intermediate steps, while ours address the mistake and achieves a correct answer.}}
    \label{fig:tf_case_replay}
    \vspace{-2mm}
\end{figure*}

\noindent {A few recent studies suggest long chain-of-thought induced by RLVR is not always beneficial and bring overthinking on simple problems~\cite{shu2025sail,wang2025skywork} Considering Activation Replay are softly reinjecting representations from instruct LMMs to RLVR post-trained counterparts, it is worth to discuss if such approach works by alleviating overthinking. Our simple conclusion is, this is not always the case. We compare a few responses for comparison purpose, one of them is presented in Figure~\ref{fig:tf_case_replay}. While post-trained LMM respond with length almost equal to that from Activation Replay, it fails to answer correctly. In this case, Activation Replay corrects its intermediate reasoning steps by only manipulating input contexts. \purple{More cases are given in Appendix}.}

\section{Related Works}
\label{sec:related_works}

\subsection{Large Multimodal Models}

Driven by the success from Large Language Models (LLMs)~\citep{touvron2023llama,touvron2023llama2,yang2025qwen3,achiam2023gpt4}, Large Multimodal Models (LMMs) evolve rapidly and empower LLMs with the ability to understand images~\citep{liu2023llava}, videos~\citep{damonlpsg2024videollama2,lin2023videollava}, or 3D scenes~\citep{zhu2024llava3d}. Pioneering studies~\citep{liu2023llava} achieve this in a straightforward way, by connecting a vision encoder~\citep{radford2021clip,zhai2023siglip} to LLM by linear projection~\citep{li2024llavaov} or Q-Former~\citep{dai2023instructblip}. Follow-up papers have primarily been focusing on collecting high-quality instruction data and adapting to diverse applications, while reasoning capability that requires step-by-step verification of LMMs are less explored. 

% Follow-up studies have mainly contributed high-quality instruction data, and visual features that can adapt to various applications such as native resolution, video understanding or even robotic control. Despite this, the reasoning scenarios that requires long-chain step-by-step verification are still challenging to LMMs. 

\subsection{Multimodal Reinforcement Learning}

Reinforcement Learning has evolutionized Large Language Models and enpowered them with strong reasoning capability thanks to Group Relative Policy Optimization (GRPO)~\citep{shao2024grpo}. Inspired by the recent success~\citep{yu2025dapo,zheng2025gspo} that incentivizes the reasoning capability from LLMs, a few recent studies have explored replicating the success to multimodal inputs~\citep{zhang2025r1,meng2025mmeureka,luo2025ursa,chen2025revisualr1,wan2025srpo,wang2025vlrethinker} to address multimodal reasoning scenariosm such as geometry~\citep{lu2021geometry3k}, chart interpretation~\citep{masry2022chartqa}, or long video reasoning~\citep{chen2025scaling,feng2025videor1,wang2025videothinker}, while more recently GRPO incentivizes thinking with image~\citep{wu2024vstar,zheng2025deepeyes,zhu2025activeo3,xu2025visualplan}. Some of them explore effective training recipes, including long chain-of-thought verifiable data construction~\citep{leng2025mmr1} or stages~\cite{chen2025revisualr1}. Few studies investigate the inner mechanisms for LMM reasoning behavior while our study falls into such category.

\subsection{Logit Lens}

Logit Lens~\citep{nanda2021logitlens} has been validated effective as a powerful interpretability tool, where it interprets the inner workings of LLMs and show that LLMs progressively transfer input tokens from output tokens along layers. Lately, it has also served as an effective approach to address negative effects in LMMs, such as object hallucination~\citep{zou2024memvr,jiang2025vlinterp,neo2025towards}. Our findings involve further explorations of input intermediate activations, with \textit{logit lens} as a tool, and an implicit approach to regularize low-entropy activations at test time.

\section{Conclusion}

% \blue{Recently, reinforcement learning has been successful in incentivizing reasoning capability in Large Multimodal Models thanks to Group Relative Policy Optimization. Nevertheless, the inner mechanisms on how GRPO affects LMMs are under exploration. We begin with exploring the hidden uncertainty of LMMs representations via logit lens and observe that GRPO enhances uncertainty for low-entropy tokens, a phenomenon consistent across multiple reasoning-enhanced LMMs. We conjecture that post-training introduces uncertainty for these inner tokens, which hinders LMMs reasoning behaviors. Based on these findings, we propose Activation Replay, a simple training-free intervention approach that boosts mathematical reasoning models with no extra inference cost. By simply replacing low-entropy tokens from pre-RLVR models to intervene GRPO inference, Activation Replay can boost mathmatical reasoning across multiple models and evaluations. We implement and compare several design choices, including replay of only low-entropy or high-entropy tokens and observe that low-entropy tokens are decisive. We evaluated our designs on three recent reasoning LMMs and six reason-demanding benchmarks and show that they can improve LMMs without any training. Our findings also indicate that Activation Replay can improve perception accuracy of GRPO enhanced models. Finally, we train GRPO with regularization on low-entropy prefilling tokens with two controlled setups, hoping to shed lights on.}

Reinforcement Learning with Verifiable Rewards (RLVR) has been a popular post-training paradigm in incentivizing reasoning capability in Large Multimodal Models (LMMs). We explore the inner workings on multiple existing post-trained LMMs, about how they are affected after RLVR. Especially, our findings suggest that RLVR shifts low-entropy input activations unexpectedly, which hinders better reasoning behaviors across multiple scenarios. We perform two controlled study to further explore the associations between the RLVR effects on low-entropy activations and LMM reasoning. We propose Activation Replay, a training-free approach that is effective in multiple reasoning scenarios, including mathematics, knowledge, agentic, and video reasoning, demonstrating the effectiveness of our design. We show that such approach also benefits Pass@K, and perform multiple ablations to justify our design choice.

\section*{Acknowledgement}
This study is funded by the Ministry of Education Singapore, under
the Tier-2 project scheme with project number MOET2EP20123-0003.\\

{
    \small
    \bibliographystyle{ieeenat_fullname}
    \bibliography{main}
}

\appendix
% \clearpage
% \setcounter{page}{1}

\clearpage
\setcounter{page}{1}
\maketitlesupplementary

\setcounter{table}{0}
\setcounter{figure}{0}
\setcounter{section}{0}
\setcounter{equation}{0}

\section{Additional Time Costs}
% \begin{table}[!ht]
%     \centering
%     \caption{Additional Time Costs with Activation Replay.}
%     \label{tab:tf_timecost}
%     \begin{small}
%     \renewcommand{\arraystretch}{1.0}
%         \begin{tabularx}{\linewidth}{p{0.32\linewidth}*{1}{>{\centering\arraybackslash}X}*{4}{>{\centering\arraybackslash}X}}
%         \toprule
%         \multicolumn{5}{c}{\textbf{Math}} \\
%         \midrule
%         \textbf{Model} & \textbf{ME} & \textbf{MN} & \textbf{WM} & \textbf{LV} \\
%         \midrule
%         MMR1-Math~\cite{leng2025mmr1} & \\
%         MM-Eureka~\cite{meng2025mmeureka} &  &  &  &  \\
%         VL-Rethinker~\cite{wang2025vlrethinker} \\
%         MM-Eureka$_{32B}$~\cite{meng2025mmeureka} &  &  &  &  \\
%         \midrule
%         \multicolumn{5}{c}{\textbf{Agent}} \\
%         \midrule
%         \textbf{Model} & \textbf{H4} & \textbf{H8} & \textbf{VM} & \textbf{VH} \\
%         \midrule
%         DeepEyes~\cite{zheng2025deepeyes} &   &   &   &   \\
%         \midrule
%         \multicolumn{5}{c}{\textbf{Video}} \\
%         \midrule
%         \textbf{Model} & \textbf{VU} & \textbf{ML} & \textbf{MV} & \textbf{VH} \\
%         \midrule
%         Video-R1~\cite{feng2025videor1} &   &   &   &   \\
%         \bottomrule
%         \end{tabularx}
%     \end{small}
% \end{table}

\begin{table}[!htbp]
    \centering
    \caption{Additional Time Costs with Activation Replay.}
    \label{tab:tf_timecost}
    \begin{small}
    \renewcommand{\arraystretch}{1.0}
        \begin{tabularx}{\linewidth}{p{0.3\linewidth}*{4}{>{\centering\arraybackslash}X}}
        \toprule
        \multirow{2.5}{*}{\textbf{}} & 
        \multicolumn{2}{c}{\textbf{MMR1-Math}} & 
        \multicolumn{2}{c}{\textbf{DeepEyes}} \\
        \cmidrule(lr){2-3} \cmidrule(lr){4-5}
        & ME & MN & H4 & H8 \\
        \midrule
        Relative Time Cost & 0.77$\times$ & 1.42$\times$ & 1.05$\times$ & 0.41$\times$ \\
        \bottomrule
        \end{tabularx}
    \end{small}
\end{table}
\begin{table}[!htbp]
    \centering
    \caption{More Pass@K results. In this table, MV and MA are short for MathVision$_{mini}$ and MathVista, respectively.}
    \label{tab:tf_passk}
    \begin{small}
    \renewcommand{\arraystretch}{1.0}
        \begin{tabularx}{\linewidth}{p{0.3\linewidth}*{4}{>{\raggedright\arraybackslash}X}}
        \toprule
        \multirow{2}{*}{\textbf{Model}} & 
        \multicolumn{2}{c}{\textbf{MV}} & 
        \multicolumn{2}{c}{\textbf{MA}} \\
        \cmidrule(lr){2-3} \cmidrule(lr){4-5}
        & @8 & @16 & @2 & @4 \\
        \midrule
        MMR1-Math~\cite{leng2025mmr1} & 62.8 & 71.0 & 77.0 & 81.3 \\
        \rowcolor{gray!15}+ \textit{replay} & 65.1~\tabup{2.3} & 72.4~\tabup{1.4} & 79.1~\tabup{2.1} & 82.3~\tabup{1.0} \\
        \midrule
        MM-Eureka~\cite{meng2025mmeureka} & 57.2 & 63.8 & 77.3 & 80.3 \\
        \rowcolor{gray!15}+ \textit{replay} & 59.5~\tabup{2.3} & 67.1~\tabup{3.3} & 77.3~\tabup{0.0} & 81.0~\tabup{0.7} \\
        \midrule
        VL-Rethinker~\cite{wang2025vlrethinker} & 67.8 & 76.9 & 77.9 & 83.6 \\
        \rowcolor{gray!15}+ \textit{replay} & 68.8~\tabup{1.0} & 78.6~\tabup{1.7} & 77.4~\tabdown{0.5} & 83.8~\tabup{0.2} \\
        \midrule
        VLAA-Thinker~\cite{chen2025vlaa} & 72.7 & 84.5 & 75.8 & 82.9 \\
        \rowcolor{gray!15}+ \textit{replay} & 74.8~\tabup{2.1} & 89.5~\tabup{5.0} & 77.2~\tabup{1.3} & 83.6~\tabup{0.7} \\
        \bottomrule
        \end{tabularx}
    \end{small}
\end{table}

\noindent \textbf{Details}. Note that Activation Replay brings addtional compute during test time, we provide a comparison over two tasks to clarify this, including MMR1-Math-v0~\cite{leng2025mmr1} over two datasets (MathVerse and MathVision~\cite{zhang2024mathverse,wang2024mathvision}) as mathematical testbeds, and DeepEyes~\cite{zheng2025deepeyes} over two subsets of HRBench~\cite{wang2024hrbench} as agentic testbeds. Note that agentic reasoning typically involves much longer reasoning traces than mathematical tasks. We also point out that our design only involve modulations of input contexts, while decoding is the same as baselines. For timecosts, baselines involve prefilling and decoding, while Activation Replay involves prefilling, decoding and test-time input context manipulations. 

\noindent \textbf{Discussions}. We find that the overall elapsed time for learnable token manipulations is around 0.15$\times$, especially considering those reasoning LMMs that output long chain-of-thought traces, the relatively elapsed time could be shorter. We observe in some cases Activation Replay is even comparably faster in overall (\textit{e.g.}, DeepEyes on HRBench 8K, as in Table~\ref{tab:tf_timecost}). For o3-like agentic LMMs characterized by multi-turn visual searching, our approach achieves slightly better performance (75.9 \textit{v.s.} 75.4) with less turns.

\begin{table}[!ht]
    \centering
    \caption{Activation Replay with Low (L) \textit{v.s.} High-Entropy Activations (H).}
    \label{tab:replay_bp_low_vs_high}
    \begin{small}
    \renewcommand{\arraystretch}{1.0}
        \begin{tabularx}{\linewidth}{p{0.25\linewidth}*{1}{>{\raggedright\arraybackslash}X}*{2}{>{\raggedright\arraybackslash}X}}
        \toprule
        \textbf{Model} &  & \textbf{MN} & \textbf{LV} \\
        \midrule
        \multirow{4}{*}{MM-Eureka} & B & 25.5 & 35.6 \\
        & R & 30.6 & 49.2 \\
        & L & 31.5~\tabup{0.9} & 51.0~\tabup{1.8} \\
        & H & 29.9~\tabdown{0.7} & 49.0~\tabdown{0.2} \\
        \bottomrule
        \end{tabularx}
    \end{small}
\end{table}

% \begin{table}[!ht]
%     \centering
%     \caption{Replay with Low \textit{v.s.} High-Entropy Activations.}
%     \label{tab:replay_bp_low_vs_high}
%     \begin{small}
%     \renewcommand{\arraystretch}{1.0}
%         \begin{tabularx}{\linewidth}{p{0.25\linewidth}*{1}{>{\raggedright\arraybackslash}X}*{3}{>{\raggedright\arraybackslash}X}}
%         \toprule
%         \textbf{Model} &  & \textbf{ME} & \textbf{MN} & \textbf{LV} \\
%         \midrule
%         \multirow{4}{*}{MM-Eureka} & B & 41.1 & 25.5 & 35.6 \\
%         & R & 45.1 & 30.6 & 49.2 \\
%         & L & 47.7~\tabup{2.6} & 31.5~\tabup{0.9} & 51.0~\tabup{1.8} \\
%         & H & & 29.9~\tabdown{0.7} & 49.0~\tabdown{0.2} \\
%         \midrule
%         \multirow{4}{*}{VL-Rethinker} & B & 41.1 & 25.5 & 35.6 \\
%         & R & 47.0 & 30.3 & 46.1 \\
%         & L & 49.2~\tabup{2.2} & 33.2~\tabup{2.9} & 49.7~\tabup{3.8} \\
%         & H & \\
%         \bottomrule
%         \end{tabularx}
%     \end{small}
% \end{table}
\begin{table}[!ht]
    \centering
    \caption{Activation Replay with Intervention (V) \textit{v.s.} Leaarnable Token Manipulation (M).}
    \label{tab:replay_iv_vs_bp}
    \begin{small}
    \renewcommand{\arraystretch}{1.0}
        \begin{tabularx}{\linewidth}{p{0.25\linewidth}*{1}{>{\raggedright\arraybackslash}X}*{3}{>{\raggedright\arraybackslash}X}}
        \toprule
        \textbf{Model} &  & \textbf{ME} & \textbf{MN} & \textbf{LV} \\
        \midrule
        \multirow{4}{*}{MM-Eureka} & B & 41.1 & 25.5 & 35.6 \\
        & R & 45.1 & 30.6 & 49.2  \\
        & V & 45.1~\tabup{0.0} & 31.6~\tabup{1.0} & 50.5~\tabup{1.3} \\
        & M & 47.7~\tabup{2.6} & 31.5~\tabup{0.9} & 51.0~\tabup{1.8} \\
        \midrule
        \multirow{4}{*}{VL-Rethinker} & B & 41.1 & 25.5 & 35.6 \\
        & R & 47.0 & 30.3 & 46.1 \\
        & V & 47.5~\tabup{0.5} & 33.5~\tabup{3.2} & 47.6~\tabup{1.5} \\
        & M & 49.2~\tabup{2.2} & 33.2~\tabup{2.9} & 49.7~\tabup{3.8} \\
        \bottomrule
        \end{tabularx}
    \end{small}
\end{table}

\begin{table}[!ht]
    \centering
    \caption{Activation Replay with Static (S) \textit{v.s.} Dynamic Thresholding (D, \textit{ours}).}
    \label{tab:replay_st_vs_dy}
    \begin{small}
    \renewcommand{\arraystretch}{1.0}
        \begin{tabularx}{\linewidth}{p{0.25\linewidth}*{1}{>{\raggedright\arraybackslash}X}*{3}{>{\raggedright\arraybackslash}X}}
        \toprule
        \textbf{Model} &  & \textbf{ME} & \textbf{MN} & \textbf{LV} \\
        \midrule
        \multirow{4}{*}{MM-Eureka} & B & 41.1 & 25.5 & 35.6 \\
        & R & 45.1 & 30.6 & 49.2 \\
        & S & 46.1~\tabup{1.0} & 32.9~\tabup{2.3} & 47.9~\tabdown{1.3} \\
         & \cellcolor{gray!15}D & \cellcolor{gray!15}47.7~\tabup{2.6} & \cellcolor{gray!15}31.5~\tabup{0.9} & \cellcolor{gray!15}51.0~\tabup{1.8} \\
        \midrule
        \multirow{4}{*}{VL-Rethinker} & B & 41.1 & 25.5 & 35.6 \\
        & R & 47.0 & 30.3 & 46.1 \\
        & S & 45.7~\tabdown{1.3} & 32.2~\tabup{1.9} & 48.5~\tabup{2.4} \\
         & \cellcolor{gray!15}D & \cellcolor{gray!15}49.2~\tabup{2.2} & \cellcolor{gray!15}33.2~\tabup{2.9} & \cellcolor{gray!15}49.7~\tabup{3.8} \\
        \bottomrule
    \end{tabularx}
    \end{small}
\end{table}

\section{Pass@K}

\noindent We provide details and more Pass@K results here, as mentioned in our manuscript. 

\noindent \textbf{Detail}. For MathVision$_{mini}$~\cite{wang2024mathvision}, we present Pass@8 and Pass@16 and for MathVista~\cite{lu2023mathvista}, we compare Pass@2 and Pass@4 instead, due to the resource limitations. We set temperature to 1.0 for all RLVR LMMs to sample responses. 

\noindent \textbf{Result}. The result is provided in Table~\ref{tab:tf_passk}. Apart from results presented in our manuscript, we present results of VLAA-Thinker~\cite{chen2025vlaa} here. The performance gains from Activation Replay mostly hold.

\begin{figure*}[!ht]
    \centering
    \includegraphics[width=0.95\linewidth]{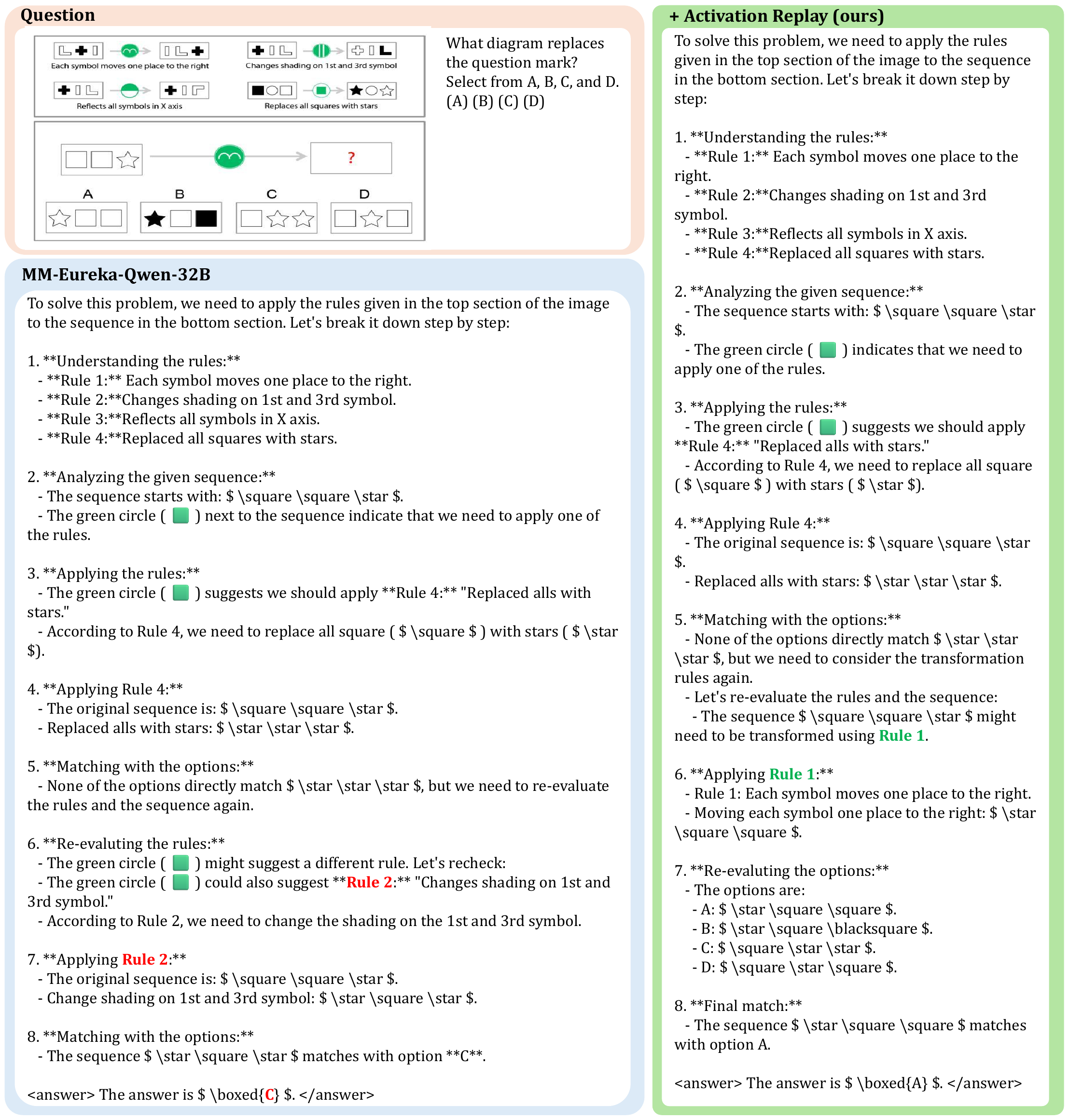}
    \caption{Case Study on Mathematical Reasoning.}
    \label{fig:math_case}
\end{figure*}
\section{Comparisons}

\subsection{Replay with High-Entropy Activations}

We apply the proposed learnable token manipulation strategy to regulate high-entropy activations instead. The results are consistent with that of intervention study from Table~\ref{tab:tf_intervention_study} (manuscript). As presented in Table~\ref{tab:replay_bp_low_vs_high}, replaying high-entropy activations, the strategy brings performance drop instead.

\subsection{Intervention}

We collect results of two replay strategies, namely direct intervention and learnable token manipulation, and present them in Table~\ref{tab:replay_iv_vs_bp}, In some cases, direct intervention are comparable to the learnable token manipulation (\textit{e.g.}, MathVision). Overall, we find that learnable token manipulation brings better performance gains.

\begin{figure*}[!ht]
    \centering
    \includegraphics[width=\linewidth]{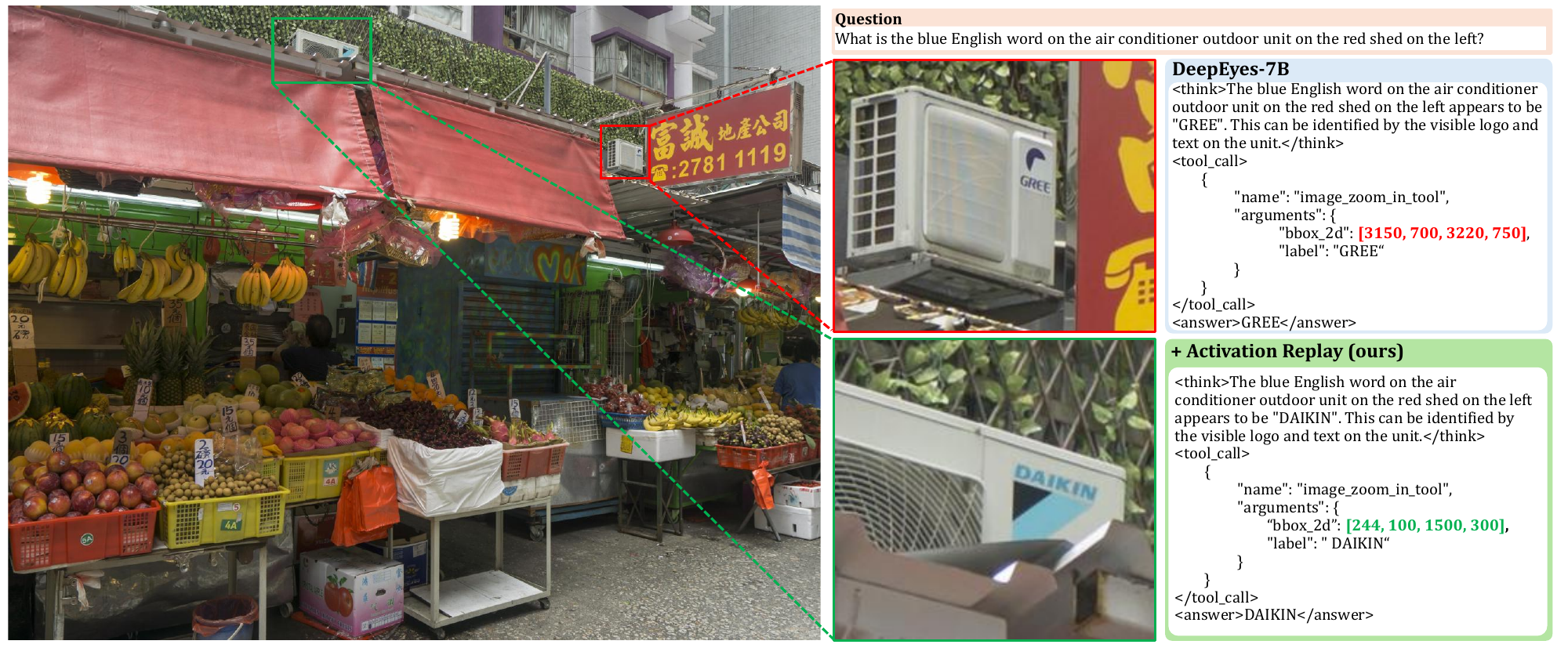}
    \caption{\textbf{False Tool Call}. Case Study of Multi-Turn o3-Like Agent~\cite{zheng2025deepeyes}.}
    \label{fig:agent_case_2}
\end{figure*}

\begin{figure*}[!ht]
    \centering
    \includegraphics[width=\linewidth]{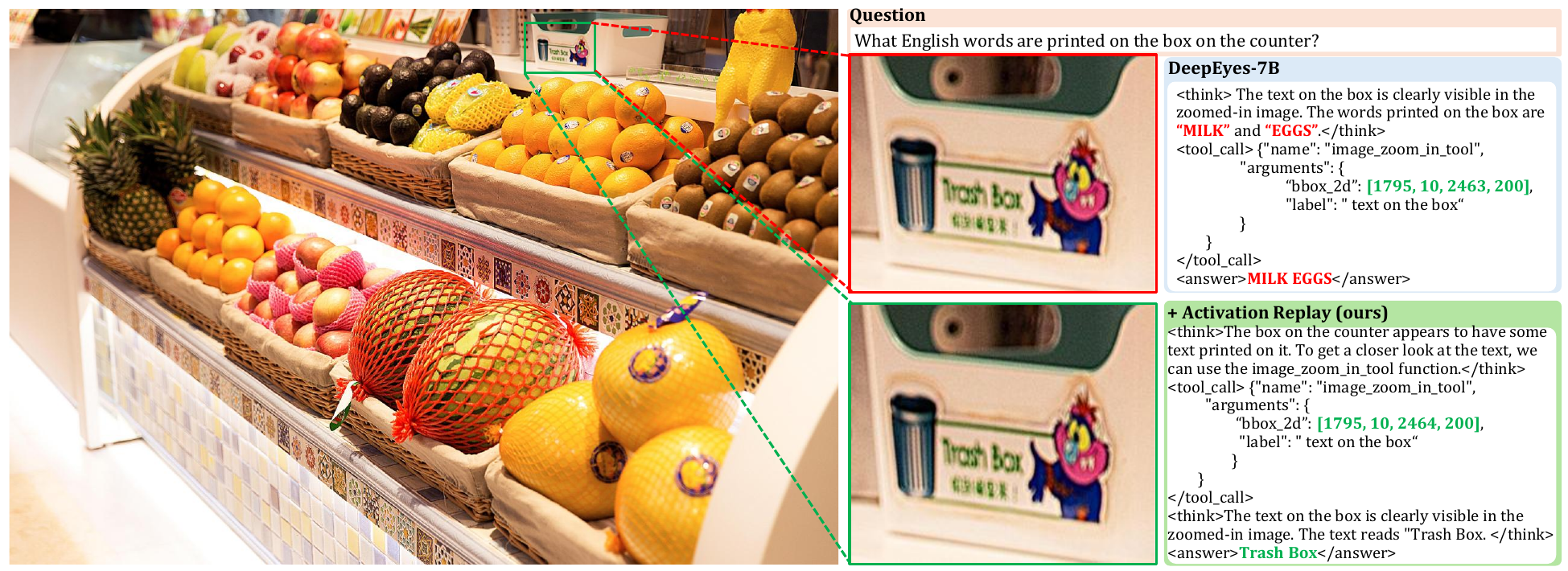}
    \caption{\textbf{False Recognition}. Case Study of Multi-Turn o3-Like Agent~\cite{zheng2025deepeyes}.}
    \label{fig:agent_case_3}
\end{figure*}
\subsection{Thresholding}

As mentioned in manuscript, we try a static thresholding to differentiate low-entropy activations from high-entropy ones. The static threshold is pre-computed by a validation set. As presented in Table~\ref{tab:replay_st_vs_dy}, although with more test-time compute, we find that the static strategy brings comparable performance gains (sometimes even performance drop), as compared to results from dynamic strategy.

% \input{sec/Suppl/7_4_low_entropy_activations}
% \section{VisionReasoner}

% \input{tab/tf_visionreasoner}
% \input{sec/Suppl/7_6_replay_coldstart}
% \section{Related Works}

\section{Case Study}

\subsection{Math}

We show a case study when applying Activation Replay to MM-Eureka-Qwen-32B~\cite{meng2025mmeureka} in Figure~\ref{fig:math_case}. Notice that despite both inferences start with apply Rule 4 mistakenly. While with re-evaluating, the RLVR LMM again apply another wrong rule (Rule 2), while with Activation Replay, the LMM successfully applies Rule 1, and answer correctly.

\subsection{Agent}

We also show two case studies to present how Activation Replay affects o3-like agentic reasoning~\cite{zheng2025deepeyes}. 

\noindent \textbf{Localisation}. One type of error correction is, RLVR LMM locates wrongly with tools. As in Figure~\ref{fig:agent_case_2}, RLVR LMM falsely pinpoints the air conditioner outdoor unit on the red shed on the \textit{right}, leading to the final false recognition (``GREE''). In comparison, Activation Replay correctly pinpoints and zoom in the region of ``air conditioner on the left'' and correctly recognizes the region. 
\begin{figure*}[!ht]
    \centering
    \includegraphics[width=\linewidth]{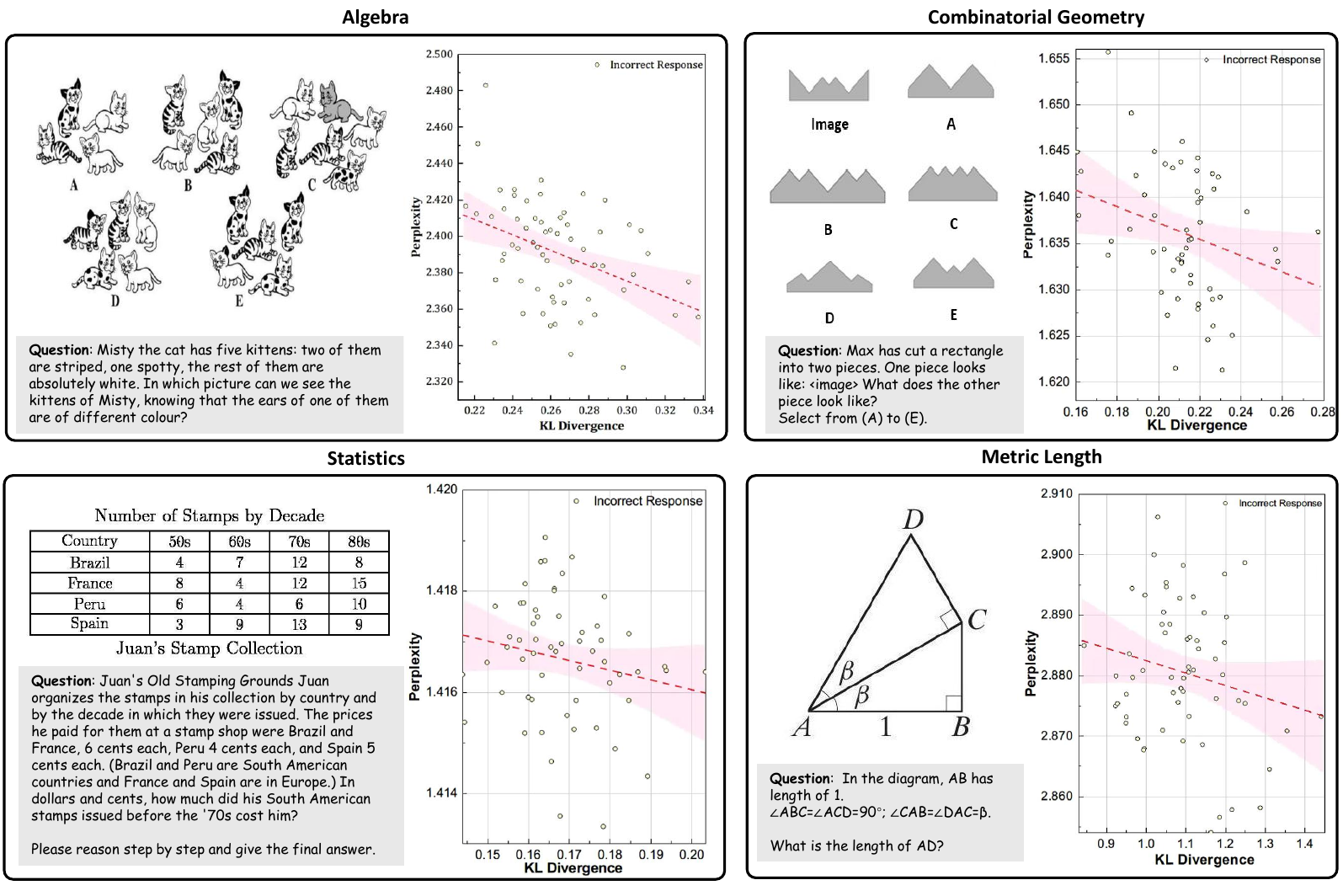}
    \caption{Perplexity Shifts on Incorrect Responses.}
    \label{fig:supp_perturbation}
\end{figure*}

\noindent \textbf{Recognition}. Another type is recognition-level error with correct tool call. As given in Figure~\ref{fig:agent_case_3}, despite RLVR LMM zooms in the correct region, the recognition is false (``MILK EGG''), while with Activation Replay, recognition is correct (``Trash Box'').

\subsection{Perturbation Study}

As mentioned in our manuscript, we also provide the perplexity shifts of incorrect responses in Figure~\ref{fig:supp_perturbation}. Note that as KL divergence of low-entropy logits increase, the perplexity of incorrect responses decreases, which is opposite to that observed for correct responses. Since Activation Replay improves Pass@1 in multiple tasks, we conjecture that this might indicate a potential cause for why our approach take effects. We leave this for future work. 
% \subsection{Video}

% \input{fig/Suppl/fig_video}

% \subsection{Segmentation}

% \input{fig/Suppl/fig_seg}
% \section{Prompt}


\end{document}